\def\year{2020}\relax
\documentclass[letterpaper]{article}
\usepackage{aaai20}
\usepackage{times}
\usepackage{helvet}
\usepackage{courier}
\usepackage[hyphens]{url}
\usepackage{graphicx}
\usepackage{graphicx}  %
\usepackage{soul}
\usepackage{booktabs}
\usepackage{times}
\usepackage{epsfig}
\usepackage{amsmath}
\usepackage{amssymb}
\usepackage{mathabx}
\usepackage{subcaption}
\usepackage{color}
\usepackage{xspace}
\usepackage{natbib}
\usepackage{url}
\usepackage{soul}
\usepackage{siunitx}
\usepackage{tikz}

\def\streetlang{StreetNav\xspace}
\def\single{{\sc Step-by-Step}\xspace}
\def\rawsingle{Step-by-Step\xspace}
\def\fixedswitcher{{\sc List + Incremental Reward}\xspace}
\def\rawfixedswitcher{List + Incremental Reward\xspace}
\def\fulltask{{\sc List + Goal Reward}\xspace}
\def\rawfulltask{List + Goal Reward\xspace}

\def\enhancedvanilla{AllConcat\xspace}
\def\summationvanilla{AllSum\xspace}
\def\hafs{Hard-A\xspace}
\def\safs{Soft-A\xspace}

\def\noinstr{NoDir\xspace}
\def\nosignal{NoSignal\xspace}
\def\notext{NoText\xspace}
\def\nothumb{NoThumb\xspace}
\def\teacher{Teacher\xspace}
\def\student{Student\xspace}
\newcommand{\variance}[1]{\raisebox{.25ex}{\scriptsize$ \pm #1$}}

\definecolor{kmhcolour}{RGB}{124, 24, 24}
\definecolor{maticolour}{RGB}{0,0,125}
\definecolor{piotrcolour}{RGB}{0,125,0}
\definecolor{rhcolour}{RGB}{255,0,255}

\newcommand{\bs}[1]{\mathbf{#1}}

\DeclareMathOperator*{\argmax}{arg\,max}

\def\checkmark{\tikz\fill[scale=0.4](0,.35) -- (.25,0) -- (1,.7) -- (.25,.15) -- cycle;}

\title{Learning to Follow Directions in Street View}
\author{Karl Moritz Hermann\thanks{Authors contributed equally.}, Mateusz Malinowski\footnotemark[1], Piotr Mirowski\footnotemark[1], \\
\bf \Large Andr\'{a}s B\'{a}nki-Horv\'{a}th, Keith Anderson, Raia Hadsell \\
DeepMind
}

\begin{document}
\maketitle

\begin{abstract}
\begin{quote}
Navigating and understanding the real world remains a key challenge in machine learning and inspires a great variety of research in areas such as language grounding, planning, navigation and computer vision.
We propose an instruction-following task that requires all of the above, and which combines the practicality of simulated environments with the challenges of ambiguous, noisy real world data.
\streetlang is built on top of Google Street View and provides visually accurate environments representing real places. Agents are given driving instructions which they must learn to interpret in order to successfully navigate in this environment.
Since humans equipped with driving instructions can readily navigate in previously unseen cities, we set a high bar and test our trained agents for similar cognitive capabilities.
Although deep reinforcement learning (RL) methods are frequently evaluated only on data that closely follow the training distribution, our dataset extends to multiple cities and has a clean train/test separation. This allows for thorough testing of generalisation ability.
This paper presents the \streetlang environment and tasks, models that establish strong baselines, and extensive analysis of the task and the trained agents.
\end{quote}
\end{abstract}

\section{Introduction}
\label{introduction}

{\it How do you get to Carnegie Hall?
\\
\-\quad --- Practice, practice, practice...}

The joke hits home for musicians and performers, but the rest of us expect actual directions. For humans, asking for directions and subsequently following those directions to successfully negotiate a new and unfamiliar environment comes naturally.
Unfortunately, transferring the experience of artificial agents from known to unknown environments remains a key obstacle in deep reinforcement learning (RL).
For humans, transfer is frequently achieved through the common medium of language. This is particularly noticeable in various navigational tasks, where access to textual directions can greatly simplify the challenge of traversing a new city.
This is made possible by our ability to integrate visual information and language and to use this to inform our actions in an inherently ambiguous world.

Recent progress in the development of RL agents that can act in increasingly more sophisticated environments~\citep{anderson2018vision,hill2017understanding,mirowski2018learning}
supports the hope that such an understanding might be possible to develop within virtual agents in the near future.
That said, the question of transfer and grounding requires more realistic circumstances in order to be fully investigated. While proofs of concept may be demonstrated in simulated, virtual environments, it is important to consider the transfer problem also on real-world data at scale, where training and test environments may radically differ in many ways, such as visual appearance or topology of the map.

\begin{figure}
\centering
\includegraphics[width=\linewidth]{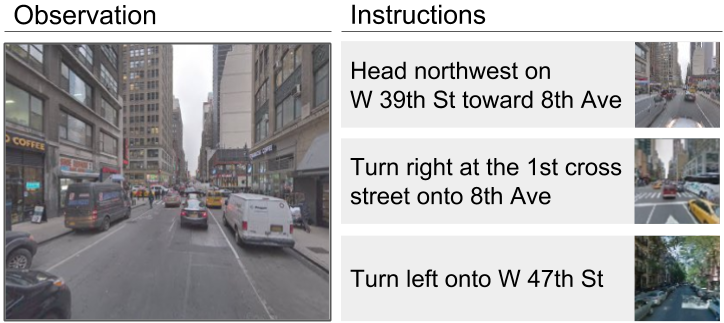} %
\caption{A route from our dataset with visual observation, instruction text and thumbnails. The agent must learn to interpret the text and thumbnails and navigate to the goal location.}
\label{fig:streetground_task}
\end{figure}

We propose a challenging new suite of RL environments termed \streetlang, based on Google Street View, which consists of natural images of real-world places with realistic connections between them. These environments are packaged together with Google Maps-based driving instructions to allow for a number of tasks that resemble the human experience of following directions to navigate a city and reach a destination (see Figure~\ref{fig:streetground_task}).

Successful agents need to integrate inputs coming from various sources that lead to optimal navigational decisions under realistic ambiguities, and then apply those behaviours when evaluated in previously unseen areas. \streetlang provides a natural separation between different neighbourhoods and cities so that such transfer can be evaluated under increasingly challenging circumstances. Concretely, we train agents in New York City and then evaluate them in an unseen region of the city, in an altogether new city (Pittsburgh), as well as on larger numbers of instructions than encountered during training. We describe a number of approaches that establish strong baselines on this problem.

\section{Related Work}
\paragraph{Reinforcement Learning for Navigation}
End-to-end RL-based approaches to navigation
jointly learn a representation of the environment (mapping) together with a suitable sequence of actions (planning). These research efforts have utilised synthetic 3D environments such as
VizDoom~\citep{kempka2016vizdoom},
DeepMind Lab~\citep{beattie2016deepmind},
HoME~\citep{brodeur2017home},
House 3D~\citep{wu2018building},
Chalet~\citep{yan2018chalet},
or AI2-THOR~\citep{kolve2017ai2}.
The challenge of generalisation to unseen test scenarios has been highlighted in~\citet{dhiman2018critical},
and partially addressed by generating maps of the environment~\citep{parisotto2018global,zhang2017neural}.

More visually realistic environments such as Matterport Room-to-Room~\citep{chang2017matterport3d}, AdobeIndoorNav~\citep{mo2018adobeindoornav}, Stanford 2D-3D-S~\citep{armeni_cvpr16}, ScanNet~\citep{dai2017scannet}, Gibson Env~\citep{xia2018gibson}, and MINOS~\citep{savva2017minos} have recently introduced to represent indoor scenes, some augmented with navigational instructions. Notably, \citet{anderson2018vision} have trained agents with supervised `student / teacher forcing' (requiring privileged access to ground-truth actions at training time) to navigate in virtual houses. We implement this method as a baseline for comparison.
\citet{mirowski2018learning} have trained deep RL agents in large-scale environments based on Google Street View and studied the transfer of navigation skills by doing limited, modular retraining in new cities, with optional adaption using aerial images \citep{li2019cross}.

\paragraph{Language, Perception, and Actions}
Humans acquire basic linguistic concepts through communication about, and interaction within their physical environment, e.g. by assigning words to visual observations~\citep{gopnik1984semantic,hoff2013language}.
The challenge of associating symbols to observations is often referred to as the symbol grounding problem~\citep{harnad1990symbol}; it has been studied using symbolic approaches such as semantic parsers~\citep{tellex2011understanding,krishnamurthy2013jointly,matuszek2012joint,malinowski2014multi},
and, more recently, deep learning~\citep{kong2014you,malinowski2018learning,rohrbach2017generating,johnson2016densecap,park2018multimodal,teney2017tips}.
Grounding has also been studied in the context of actions, but with the most of the focus on synthetic, small-scale environments~\citep{hermann2017grounded,hill2017understanding,chaplot2017gated,shah2018follownet,yan2018chalet,das2018embodied}.

In terms of more realistic environments,
\citet{anderson2018vision} and \citet{fried2018speaker}
consider the problem of following textual instructions in Matterport 3D.
\citet{de2018talk} use navigation instructions and New York imagery, but rely on categorical annotation of nearby landmarks rather than visual observations and use a smaller dataset of 500 panoramas (ours is two orders of magnitude larger).
Recently, \citet{cirikfollowing} and \citet{chen2018touchdown} have also proposed larger datasets of driving instructions grounded in Street View imagery.
Our work shares a similar motivation to \citep{chen2018touchdown}, with key differences being that their agents observed both their heading and by how much to turn to reach the next street/edge in the graph, whereas ours need to learn what is a traversable direction purely from vision and multiple instructions.
See Table~\ref{tab:comparison} for a comparison.

\section{The \streetlang Suite}
\label{sec:our_suite}
\begin{table*}[tbh]
\centering
\begin{tabular}{@{}lrrllccc@{}}
\toprule
Dataset/Paper & \#routes & $\diameter$steps & Actions &Type & Nat. Lang. &  Public & Disjoint Split \\
\midrule
\streetlang & 613,000 & 125 & Discretised & Outdoor &   & \checkmark& \checkmark \\
Room-to-Room$^a$ & 7,200 & 6 & Discretised & Indoor & \checkmark  & \checkmark & \checkmark \\
Touchdown$^b$ & 9,300 & 35 & Simplified & Outdoor & \checkmark  & \checkmark &  \\
Formulaic Maps$^c$ & 74,000 & 39 & Simplified & Outdoor &   & & \checkmark \\
\bottomrule
\end{tabular}
\caption{Comparison of real world navigation datasets. a: \protect\cite{anderson2018vision}, b: \protect\cite{chen2018touchdown}, c: \protect\cite{cirikfollowing}. Steps denotes the average number of actions required to successfully complete a route. Room-to-room and \streetlang use more complex (Discretised) action space than Touchdown and Formulaic Maps (Simplified, where agents always face a valid trajectory and cannot waste an action going against a wall or sidewalk). Nat. Lang. denotes whether natural language instructions are used. Disjoint split refers to different environments that can be used in train and test times.
}
\label{tab:comparison}
\end{table*}

We design navigation environments in the \streetlang suite by extending the dataset and environment available from StreetLearn\footnote{An open-source environment built with Google Street View for navigation research: \url{http://streetlearn.cc}}
through the addition of driving instructions from Google Maps by randomly sampling start and goal positions.

We designate geographically separated training, validation and testing environments. Specifically, we reserve Lower Manhattan in New York City for training and use parts of Midtown for validation. Agents are evaluated both in-domain (a separate area of upper NYC), as well as out-of-domain (Pittsburgh). We make the described environments, data and tasks available at \url{http://streetlearn.cc}.

Each environment $\mathcal{R}$ is an undirected, connected graph $\mathcal{R} = \{\mathcal{V}, \mathcal{E}\}$, with a set of nodes $\mathcal{V}$ and connecting edges $\mathcal{E}$. Each node $v_i \in \mathcal{V}$ is a tuple containing a $360^\degree$ panoramic image $p_i$ and a location $c_i$ (given by latitude and longitude). An edge $e_{ij} \in \mathcal{E}$ means that node $v_i$ is accessible from $v_j$ and vice versa.
Each environment has a set of associated \emph{routes} which are defined by a start and goal node, and a list of textual instructions and thumbnail images which represent directions to waypoints leading from start to goal. Success is determined by the agent reaching the destination described in those instructions within the allotted time frame.
This problem requires from agents to understand the given instructions, determine which instruction is applicable at any point in time, and correctly follow it given the current context.
Dataset statistics are in Table \ref{tab:dataset}. %
\begin{table}[ht]
    \setlength\tabcolsep{3pt} %
    \centering%
    \begin{tabular}{@{}l@{ }rrrrr@{}}
    \toprule
    Environment & \#routes &
    Avg:\quad len &
    steps &
    instrs &
     $|instr|$ \\
    \midrule
    NYC train & 580,415 & 1,194 & 128 & 4.0 & 7.1 \\
    NYC valid & 10,000 & 1,184 & 125 & 3.7 & 8.1 \\
    NYC test & 10,000 & 1,180 & 123 & 3.8 & 7.9 \\
    NYC larger & 3,923 & 1,667 & 174 & 6.5 & 8.1  \\
    Pittsburgh test & 8,474 & 998 & 106 & 3.8 & 6.6  \\
    \bottomrule
    \end{tabular}
    \caption{Dataset statistics: the number of routes, their average length in meters and in environment steps, the average number of instructions per route and average number of words per instruction.}
    \label{tab:dataset}
\end{table}

\subsection{Driving Directions}
By using driving instructions from Google Maps we make a conscious trade-off between realism and scale.
The language obtained is synthetic, but it is used by millions of people to see, hear, and follow instructions every day, which we feel justifies its inclusion in this `real-world' problem setting. We believe the problem of grounding such a language is a sensible step to solve before attempting the same with natural language; similar trends can be seen in the visual question answering community~\citep{hudson2019gqa,johnson2017clevr}. Figure \ref{fig:streetground_task} shows a few examples of our driving directions, with more in the appendix or easily found by using Google Maps for directions.

\subsection{Agent Interface}
To capture pragmatics of navigation,
we model the agent's observations using Google Street View first-person view interface combined with instructions from Google Maps. We therefore mimick the experience of a user following written navigation instructions.
At each time step, the agent receives an observation and a list of instructions with matching thumbnails. Observations and thumbnails are RGB images taken from Google Street View, with the observation image representing the agent's field of view from the current location and pose. The instructions and thumbnails are drawn from the Google Directions API and describe a route between two points.
$n$ instructions are matched with ${n+1}$ thumbnails, as both the initial start and the final goal are included in the list of locations represented by a thumbnail.

RGB images are $60^\degree$ crops from the panoramic image that are scaled to 84-by-84 pixels.
We have five actions:
move forward, slow ($\pm10^\degree$), and fast rotation ($\pm30^\degree$). While moving forward, the agent takes the graph edge that is within its viewing cone and most closely aligned to the agent's current orientation; the lack of a suitable graph edge results in a NO-OP. Therefore, the first challenge for our agent is to learn to follow a street without going onto the sidewalk. The episode ends automatically if the agent reaches the last waypoint or after $1000$ time steps. This, combined with the difficult exploration problem (the agent could end up kilometres away from the goal) forces the agent to plan trajectories.

Table \ref{tab:comparison} lists differences and similarities between our and related, publicly available datasets with realistic visuals and topologies, namely \emph{Room-to-Room} \citep{anderson2018vision}, \emph{Touchdown} \citep{chen2018touchdown} and \emph{Formulaic Maps} \citep{cirikfollowing}.

\subsection{Variants of  the \streetlang Task}
To examine how an agent might learn to follow visually grounded navigation directions, we propose three task variants, with the increasing difficulty.
They share the same underlying structure, but differ in how directions are presented (step-by-step, or all at once) and whether any feedback is given to agents when reaching intermediate waypoints. %

\subsubsection{Task 1: \rawfulltask}
\label{sec:fulltask}

The \fulltask task mimics the human experience of following printed directions, with access to the complete set of instructions and referential images but without any incremental feedback as to whether one is still on the right track.
Formally, at each step the agent is given as input a list of directions $\mathbf{d} = \langle d_1,d_2,\dots d_N\rangle$ where $d_i = \{\iota_i,t^s_i,t^e_i\}$ (instruction, start and end thumbnail).\footnote{Note that $t^e_i = t^s_{i+1}$ for all $1<i<N$} $\iota_i = \langle \iota_{i,1},\iota_{i,2},\dots \iota_{i,M}\rangle$ where $\iota_{i,j}$ is a single word token. $t^s_i, t^e_i$ are thumbnails in $\mathbb{R}^{3\times 84\times 84}$. The number of directions $N$ varies per route and the number of words $M$ varies according to the instruction.

An agent begins each episode in an initial state $s_0 = \langle v_0, \theta_0\rangle$ at the start node and heading associated with the given route, and is given an RGB image $x_0$ corresponding to that state. The goal is defined as the final node of the route, $G$. The agent must generate a sequence of actions $\langle s_0,a_0,s_1,a_1,\dots s_T\rangle$, with each action $a_t$ leading to a modified state $s_{t+1}$ and corresponding observation $x_{t+1}$.

The episode ends either when a maximal number of actions is reached $T > T_{\text{max}}$, or when the agent reaches $G$. The goal reward $R_g$ is awarded if the agent reaches the final goal node $G$ associated with the given route, i.e.
 $r_t = R_g \text{ if } v_t = G$. This is a hard RL task, with a sparse reward.

\subsubsection{Task 2: \rawfixedswitcher}

This task uses the same presentation of instructions as the previous task (the full list is given at every step), however we mitigate the challenge of exploration by increasing the reward density. In addition to the goal reward, a smaller reward $R_w$ is awarded, as the input for the agent, for the successful execution of individual directions, meaning that the agent is given positive feedback when reaching any of the waypoints for the first time: %
$r_t = \langle R_w \text{ if } v_t \in V_w \land \forall_{i<t}\: v_t \neq v_i  \rangle + \langle R_g \text{ if } v_t = G \rangle$
where $V_w$ denotes the set of all the waypoints in the given route.
This formula simplifies learning of when to switch from one instruction to another.

\subsubsection{Task 3: \rawsingle}
In the simplest variant, agents are provided with a single instruction at a time, automatically switching to the next instruction whenever a waypoint it reached, similar to the human experience of following directions given by GPS.
Thereby two challenges that the agents have to solve in the previous tasks---`when to switch' and `what instruction to switch to'---are removed.
As in the \rawfixedswitcher task, smaller rewards are given as waypoints are reached.

\subsection{Training and Evaluation}

Reward shaping can be used to simplify exploration and to make learning more efficient. We use early rewards within a fixed radius at each waypoint and the goal, meaning that an agent will receive fractional rewards once it is within a certain distance ($50m$).
Reward shaping is only used for training, and never during evaluation. %
We report the percentage of goals reached by a trained agent in an environment (training, validation or test). Agents are evaluated for 1 million steps with a 1,000 step limit per route, which is sufficiently small to avoid any success by random exploration of the map. We do not consider waypoint rewards as a partial success, and we give a score 1 only if the final goal is reached.

\section{Architectures}
\label{sec:architectures}
We approach the challenge of grounded navigation with a deep reinforcement learning framework and formalise the learning problem as a Markov Decision Process (MDP). Let us consider an environment $\mathcal{R} \in \mathbb{S}$ from the suite of the environments $\mathbb{S}$. Our MDP is a tuple consisting of states that belong to the environment,
i.e. $\mathcal{S} \in \mathcal{R}$, together with the possible directions $\mathcal{D} = \{\mathbf{d}_l\}_l \in\mathcal{E}$, and possible actions $\mathcal{A}$. Each direction $\mathbf{d}$ is a sequence of instructions, and pairs of thumbnails,
i.e. $\mathbf{d} =  \langle d_1,\dots,d_n\rangle$ where $d_i = \{\iota_i,t^s_i,t^e_i\}$ (instruction, starting thumbnail, ending thumbnail). The length of $\mathbf{d}$ varies between the episodes.
Each state $s\in\mathcal{S}$ is associated with a real location, and has coordinates $c$ and observation $x$ which is an RGB image of the location. Transitions from state to state are limited to rotations and movements to new locations that are allowed by the edges in the connectivity graph. The reward function, $\mathcal{R} : \mathcal{S} \bigtimes \mathcal{D} \to \mathbb{R}$, depends on the current state and the final goal $d_g$.

Our objective, as typical in the RL setting, is to learn a policy $\pi$ that maximises the expected reward $E[\mathcal{R}]$. In this work, we use a variant of the REINFORCE~\cite{williams1992simple} advantage actor-critic algorithm $E_{\pi}\left[\sum_t \nabla_{\theta} \log \pi(a_t|s_t, \mathbf{d}; \theta) (\mathcal{R}_t - \mathcal{V}^{\pi}(s_t))\right]$, where $\mathcal{R}_t = \sum_{j=0}^{T-t} \gamma^{j}r_{t+j}$,  $[r_t]_t$ is a binary vector with one at $t$ where the final destination is achieved within the 1-sample Monte Carlo estimate of the return, $\gamma$ is a discounting factor, and $T$ is the episode length.

In the following, we describe methods that transform input signals into vector representations and are combined with a
recurrent network to learn the optimal policy $\pi^{\ast} = \argmax_\pi E_\pi[\mathcal{R}]$ via gradient descent. We also describe several baseline architectures: simple adaptations of existing RL agents as well as ablations of our proposed architectures.

\subsection{Input-to-Multimodal Representation}
We use an LSTM
to transform the text instructions into a vector. We also use CNNs
to transform RGB images, both observations and thumbnails, into vectors. Using deep learning architectures to transform raw inputs into a unified vector-based representation has been found to be very effective.
Since observations and thumbnails come from the same source, we share the weights of the CNNs. Let $\boldsymbol{\iota} = \text{LSTM}_{\theta_1}(\iota)$, $\bs{x} = \text{CNN}_{\theta_2}(x)$, $\bs{t}^s = \text{CNN}_{\theta_2}(t^s)$, and $\bs{t}^e = \text{CNN}_{\theta_2}(t^e)$ be vector representations of the instruction, observation, start-, and end-thumbnail respectively.

We use a three-layer MLP, with 256 units per layer, whose input is a concatenation of these vectors representing signals, and whose output is their merged representation.
We use this module twice, 1) to embed  instructions and the corresponding thumbnails $\bs{i} = \text{MLP}(\boldsymbol{\iota}_i, \bs{t}^s_i, \bs{t}^e_i)$, 2) to embed this output with the current observation, i.e. $\bs{p} = \text{MLP}(\bs{x}, \bs{i})$. The output of the second module is input to the policy network. This module is shown to be important in our case.

\subsection{Previous Reward and Action}
\label{sec:bla}
Optionally, we input the previous action and obtained reward to the policy network. In that case the policy should formally be written as $\pi(a_t | s_t, a_{t-1}, \mathbf{d} ; \theta)$ or $\pi(a_t | s_t, a_{t-1}, \mathcal{R}_{t-1}, \mathbf{d} ; \theta)$.
Note that adding the previous reward is only relevant when intermediate rewards are given for reaching waypoints, in which case the reward signal can be explicitly or implicitly used to determine which instruction to execute next, and that this architectural choice is unavailable in \fulltask (Section \ref{sec:fulltask}).

\begin{figure}[t]
\centering
\includegraphics[width=0.9\linewidth]{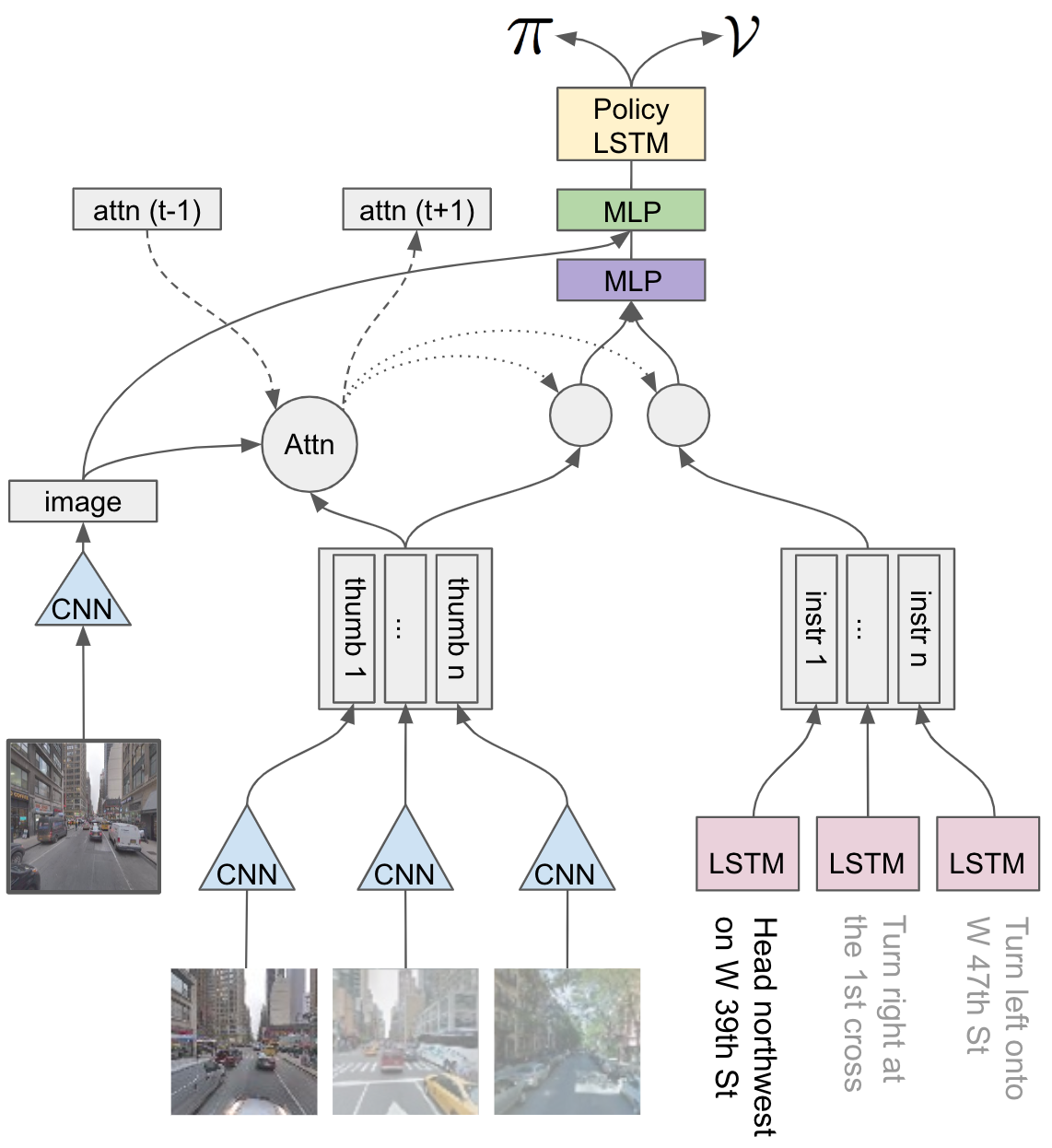}
\caption{Architecture of an agent with attention. Feature representations are computed for the observation and for all directions, using a CNN and a LSTM. The attention module  selects a thumbnail and passes it and the associated text instruction to the first multimodal MLP, whose output is concatenated with the image features and fed through the second MLP before being input to the policy LSTM. The policy LSTM outputs a policy $\pi$ and a value function $\mathcal{V}$. Colours point to components that share weights.}
\label{fig:streetground_controller}
\end{figure}

\subsection{Non-attentional Architectures}
\label{subsec:vanilla_architectures}
We introduce two architectures which are derived from the IMPALA agent and adapted to  work in this setting~\citep{espeholt2018impala}. The agent does not use attention but rather observes all the instructions and thumbnails at every time step. To accomplish this, we either concatenate the representations of all inputs (\enhancedvanilla) or first
sum over all instructions and then concatenate this summed representation with the observation (\summationvanilla), before passing the result of this operation as input to the multi-modal embedding modules.

With this type of the architecture, the agent does not explicitly decide `when to switch' or `what to switch to', but rather relies on the policy network to learn a suitable strategy to use all the instructions. When explicitly cued by the waypoint reward as an input signal, the decision when to move to the next instruction should be reasonably trivial to learn, while more problematic when trained or tested without that signal. Note that the concatenated model will not be able to transfer to larger numbers of instructions.

\subsection{Attentional Architectures}
\label{subsec:attentional_architectures}

We also consider architectures that use attention to select an instruction to follow as well as deciding whether to switch to a new instruction. We hypothesise that by factoring this out of the policy network, the agent will be capable of generalising to a larger number of instructions, while the smaller, specialised components could allow for easier training.

First, we design an agent that implements the switching logic with a hard attention mechanism, since selecting only one instruction at a time seems appropriate for the given task. Hard attention is often modelled as a discrete decision processes and as such difficult to incorporate with gradient-based optimisation.
We side-step this issue by conditioning the hard attention choice on an observation($\bs{x}$) / thumbnail ($\bs{t}_i$) similarity metric, and then selecting the most suitable instruction via a generalisation of the max-pooling operator, i.e.,
$\bs{t}_{i^\ast} = \text{argmax}_{\bs{t}_i} \left [ \text{softmax}(-||\bs{t}_i - \bs{x}||_2)\right ]$.
This results in a sub-differentiable model which can then be combined with other gradient-based components \citep{malinowski2018learning}.

We implement the `when to switch' logic as follows: when the environment signals the reaching of waypoints to the model, switching becomes a function of that. However, when this is not the case -- as in \fulltask -- we use the thumbnail-observation representation similarity to determine whether to switch:
\[
i_t =
\begin{cases}
 i^\ast, & \text{if } \text{softmax}(-||\bs{t}_{i^\ast} - \bs{x}||_2) > \tau \\
 i_{t-1}, & \text{otherwise}
\end{cases}
\]
where the threshold parameter $\tau$ is tuned on the validation set.
As this component is not trained explicitly, we can in practice train agents by manually switching at waypoint signals and only use the threshold-based switching architecture during evaluation.

Finally, we also adapted a `soft' attention mechanism from \citet{yang2016stacked} that re-weights the representations of instructions instead of selecting just one instruction. We use $h_i = W_a \tanh\left(W_x\bs{x} + W_t \bs{t}_i\right)$ to compute the unnormalised attention weights. Here, $\bs{x}, \bs{t}_i \in \mathbb{R}^{d}$ are image and i-th thumbnail representations respectively.
The normalised weights $p_i = \text{softmax}(h_i)$ are used to weight the instructions and thumbnails, i.e. $\hat{\boldsymbol{\iota}} = \sum_i p_i \boldsymbol{\iota}_i$, $\hat{\bs{t}} = \sum_i p_i \bs{t}_i$, with the resultant representation being fed to the policy module.

\subsection{Baselines and Ablations}
\label{sec:baselines_ablations}
To better understand the environment and the complexity of the problem, we have conducted experiments with various baseline agents.
\paragraph{Random and Forward}

We start with two extremely simple baselines. An agent choosing a random action at any point, and another agent always selecting the forward action. These baselines are mainly to verify that the environment is sufficiently complex to prevent success by random exploration over the available number of steps.

\paragraph{No-Directions}

We train and evaluate an agent that only takes observations $x$ as input and ignores the instructions, thus establishing a baseline agent that, presumably, will do little more than memorise the training data or perhaps discover exploitable regularities in the environment.
We compare \noinstr, which uses the waypoint reward signal when available, and \nosignal which does not. The former naturally has more strategies available to exploit the game.

\paragraph{No-Text and No-Thumbnails}

To establish the relative importance of text instructions and waypoint thumbnails we further consider two variants of the agent where one of these inputs is removed. The \nothumb agent is built on top of the no-attention architecture (Section~\ref{subsec:vanilla_architectures}), while the \notext version is based on the attentional architecture (Section~\ref{subsec:attentional_architectures}).

\paragraph{Student and Teacher Forcing on Ground-truth Labels}
In addition to our main experiments,
we also consider a simple, supervised baseline.
Here, we use multinomial regression of each predicted action from the agent's policy to the ground-truth action, similarly to~\citet{anderson2018vision}.
For every waypoint, we compute the shortest path from the current agent location, and derive the optimal action sequence (turns and forward movements) from this. In \student forcing,
the agent samples an action according to the learnt policy, whereas in \teacher forcing, the agent always executes the ground-truth action. Note that the forcing is only done during training, not evaluation.
In contrast to our main experiments with RL agents, this baseline requires access to ground-truth actions at each time step, and it may overfit more easily.

\section{Experiments and Results}

Here we describe the training and evaluation of the proposed models on  different tasks outlined in Section~\ref{sec:our_suite},
followed by analysis of the results in Section \ref{sec:analysis}.

\subsection{Experimental Setup}
\paragraph{Training, Validation, and Test}

In all experiments, four agents are trained for a maximum of 1 billion steps\footnote{We randomly sample learning rates ($\num{1E-4}\leq\lambda\leq\num{2.5e-4}$) and entropy ($\num{5e-4}\leq\sigma\leq\num{5e-3}$) for each training run.}. We use an asynchronous actor-critic framework with importance sampling weights~\citep{espeholt2018impala}. We choose the best performing agent through evaluation on the validation environment, and report mean and standard deviation over three independent runs on the test environments.

\paragraph{Curriculum Training}
As the \streetlang suite is composed of tasks of increasing complexity, we can use these as a natural curriculum. We first train on the \single task, and fine-tune the same agent by continuing to train on the \fixedswitcher task. Agents that take waypoint reward signals as input are then evaluated on the \fixedswitcher task, while those that do not are evaluated on the \fulltask task.

\paragraph{Visual and Language Features}
We train our visual encoder end-to-end together with the whole architecture using 2-layer CNNs. We choose the same visual architecture as~\citet{mirowski2018learning} for the sake of the comparison to prior work, and since this design decision is computationally efficient.
In two alternative setups we use $2048$ dimensional visual features from the second-to-last layer of a ResNet~\citep{he2016deep} pre-trained on ImageNet~\citep{russakovsky2015imagenet} or on the Places dataset~\citep{zhou2017places}.
However, we do not observe improved results with the pre-trained features. In fact, the learnt ones consistently yield better results. We assume that owing to the end-to-end training and large amount of data provided, agents can learn visual representations better tailored to the task at hand than can be achieved with the pre-trained features. We report our results with learnt visual representations only.

We encode the text instructions using a word-level LSTM.
We have experimented with learnt and pre-trained word embeddings and settled on Glove embeddings~\citep{pennington2014glove}.

\paragraph{Stochastic MDP}
To reduce the effect of memorisation, we inject stochasticity into the MDP during training. With a small probability $p=0.001$ the agent cannot execute its `move forward' action.

\begin{table}[ht]
    \setlength\tabcolsep{6pt} %
    \centering
    \begin{tabular}{@{}lrrrr@{}}
    \toprule
    Model & \multicolumn{1}{l}{Training} & \multicolumn{1}{l}{Valid.} & \multicolumn{2}{@{}c@{}}{Test} \\
    \cmidrule{4-5}
    & & & \multicolumn{1}{l}{NYC} & \multicolumn{1}{l@{}}{Pittsburgh} \\
    \midrule
    Random & 0.8\variance{0.2} & 0.8\variance{0.1} & 0.8\variance{0.2} & 1.3\variance{0.4} \\
    Forward & 0.7\variance{0.3}& 0.2\variance{0.2}& 0.9\variance{0.1} & 0.6\variance{0.1}\\
    \midrule
    \nosignal & 3.5\variance{0.2} & 1.9\variance{0.6} & 3.5\variance{0.3} & 5.2\variance{0.6}  \\
    \noinstr & 57.0\variance{1.1} & 51.6\variance{0.9} & 41.5\variance{1.2} & 15.9\variance{1.3} \\
    \notext & 84.5\variance{0.7} & 58.1\variance{0.5} & 47.1\variance{0.7} & 16.9\variance{1.4} \\
    \nothumb & 90.7\variance{0.3} & 67.3\variance{0.9} & 66.1\variance{0.9} & 38.1\variance{1.7} \\
    \midrule
    \student & 94.8\variance{0.9} & 4.6\variance{1.4} & 5.5\variance{0.9} & 0.9\variance{0.2} \\ %
    \teacher & \textbf{95.0}\variance{\textbf{0.6}} & 22.9\variance{2.7} & 23.9\variance{1.9} & 8.6\variance{0.9} \\ %
    \midrule
    All-*$^+$ & 89.6\variance{0.9} & 69.8\variance{0.4} & 69.3\variance{0.9} & 44.5\variance{1.1} \\
    Hard-A$^+$ & 83.5\variance{1.0} & \textbf{74.8}\variance{\textbf{0.6}} & \textbf{72.7}\variance{\textbf{0.5}} & \textbf{46.6}\variance{\textbf{0.8}} \\
    Soft-A$^+$ & 89.3\variance{0.2} & 67.5\variance{1.4} & 66.7\variance{1.1} & 37.2\variance{0.6} \\
    \bottomrule
    \end{tabular}
    \caption{\single (instructions are given one at a time as each waypoint is reached): Percentage of goals reached. Higher is better. $\pm$ denotes standard deviation over 3 independent runs of the agent. $^+$: Note that \enhancedvanilla and \summationvanilla are equivalent in this setup. Further, the attention components of Soft-A and Hard-A are not used here, but results differ from the All-* agents due to the additional multi-modal projections used in those models.}
    \label{tab:singletask}
\end{table}

\begin{table}[ht]
    \setlength\tabcolsep{4pt} %
    \centering
    \begin{tabular}{@{}lrrrr@{}}
    \toprule
    Model & \multicolumn{1}{l}{Training} & \multicolumn{1}{l}{Valid.} & \multicolumn{2}{@{}c@{}}{Test} \\
    \cmidrule{4-5}
    & & & \multicolumn{1}{l}{NYC} & \multicolumn{1}{l@{}}{Pittsburgh} \\
    \midrule
    \notext & 53.5\variance{1.2} & 43.5\variance{0.5} & 32.6\variance{2.1} & 15.8\variance{1.7} \\
    \nothumb & 69.7\variance{0.7} & 58.7\variance{2.1} & 52.4\variance{0.4} & \textbf{33.9}\variance{\textbf{2.2}} \\
    \midrule
    \enhancedvanilla & 64.5\variance{0.6} & \textbf{61.3}\variance{\textbf{0.9}} & \textbf{53.6}\variance{\textbf{1.1}} & \textbf{33.5}\variance{\textbf{0.2}} \\
    \summationvanilla & 59.9\variance{0.5} &  51.1\variance{1.1} & 41.6\variance{1.0} & 19.1\variance{1.4}  \\
    \summationvanilla$_\text{tuned}$ & \textbf{84.4}\variance{\textbf{0.4}} & 57.7\variance{0.8} & 48.3\variance{0.9} & 22.1\variance{0.9}  \\
    \hafs & 55.4\variance{1.8} & 51.1\variance{1.1} & 42.6\variance{0.7} & 24.0\variance{0.5}\\
    \hafs$_\text{tuned}$ & 62.5\variance{1.1} & 57.9\variance{0.5} & 42.9\variance{0.6} & 22.5\variance{2.0} \\
    \safs & 74.8\variance{0.4} & 52.2\variance{1.0} & 43.2\variance{2.2} & 23.0\variance{0.9} \\
    \safs$_\text{tuned}$ & 82.7\variance{0.1} & 57.9\variance{2.1} & 44.1\variance{1.8} & 26.6\variance{0.5}\\
    \bottomrule
    \end{tabular}
    \caption{\fixedswitcher: Percentage of goals reached. Higher is better. $\pm$ denotes standard deviation over 3 independent runs of the agent. {\it tuned} denotes agents that were first trained on \single and subsequently directly on the \fixedswitcher task.}
    \label{tab:conetask}
\end{table}

\begin{table}[ht]
    \setlength\tabcolsep{4pt} %
    \centering
    \begin{tabular}{@{}lrrrr|rrr@{}}
    \toprule
    Model & \multicolumn{7}{@{}c@{}}{Number of instructions} \\
    \cmidrule{2-8}
    & 2 & 3 & 4 & 5 & 6 & 7 & 8 \\
    \midrule
    \noinstr & 72.3 & 59.4 & 49.2 & 44.1 & 31.5 & 28.0 & 11.9\\
    \nosignal & 5.4 & 1.4 & 2.2 & 0.7 & 0.1 & 0.1 & 0.0 \\
    \midrule
    \summationvanilla & 68.0 & 57.9 & 43.8 & 37.1 & 24.0 & 20.9 & 9.5 \\
    \hafs & 66.7 & 56.7 & 48.7 & 40.4 & 29.5 & 29.1 & 9.2 \\
    \hafs$_\text{tuned}$ & 71.9 & \textbf{62.9} & 52.1 & 45.7 & \textbf{32.8} & \textbf{33.8} & \textbf{15.1} \\
    \safs & \textbf{75.4} & 62.0 & 46.7 & 41.7 & 29.7 & 26.2 & 7.3 \\
    \safs$_\text{tuned}$ & 72.4 & 62.7 & \textbf{54.8} & \textbf{46.1} & 31.0 & 28.2 & 12.3 \\
    \bottomrule
    \end{tabular}
    \caption{Comparison of results on the \fixedswitcher task with a larger number of instructions than encountered during training. Number is percentage of goals reached.}
    \label{tab:larger}
\end{table}

\begin{table}[ht]
    \setlength\tabcolsep{6pt} %
    \centering
    \begin{tabular}{@{}lrrrr@{}}
    \toprule
    Model & \multicolumn{1}{l}{Training} & \multicolumn{1}{l}{Valid.} & \multicolumn{2}{@{}c@{}}{Test} \\
    \cmidrule{4-5}
    & & & \multicolumn{1}{l}{NYC} & \multicolumn{1}{l@{}}{Pittsburgh} \\
    \midrule
    \noinstr & 3.5\variance{0.2} & 1.9\variance{0.6} & 3.5\variance{0.3} & 5.2\variance{0.6} \\
    \midrule
    \enhancedvanilla & \textbf{23.0}\variance{\textbf{0.8}} & 7.4\variance{0.2} & 11.3\variance{1.8} & 9.3\variance{0.8} \\
    \summationvanilla$_\text{step}$ & 6.7\variance{1.1} & 4.1\variance{0.3} & 5.6\variance{0.1} & 7.2\variance{0.4} \\
    \summationvanilla$_\text{list}$ & 13.2\variance{2.0} & 3.2\variance{0.3} & 6.7\variance{1.1} & 5.3\variance{1.2} \\
    \hafs$_\text{step}$ & 18.5\variance{1.3} & 13.8\variance{1.3} & \textbf{17.3}\variance{\textbf{1.2}} & \textbf{12.1}\variance{\textbf{1.1}} \\
    \hafs$_\text{list}$ & 21.9\variance{2.8} & \textbf{14.2}\variance{\textbf{0.4}} & 16.9\variance{1.2} & 10.0\variance{1.3} \\
    \bottomrule
    \end{tabular}
    \caption{\fulltask: Percentage of goals reached. Higher is better.
    $\pm$ denotes standard deviation over 3 independent runs of the agent.
    Switching thresholds for the attention agent are tuned on the validation data. \textit{step} and \textit{list} denote whether the agents were trained in the \single or \fixedswitcher setting.
    }
    \label{tab:hardtask}
\end{table}

\subsection{Results and Analysis}\label{sec:analysis}

Table~\ref{tab:singletask} presents the \single task, where `what to switch to' and `when to switch' are abstracted from the agents.
Table~\ref{tab:conetask} contains the \fixedswitcher results and contains the main ablation study of this work.
Finally, Table~\ref{tab:hardtask} shows  results on the most challenging \fulltask task.

The level of difficulty between the three task variants is apparent from the relative scores. While agents reach the goal over 50\% of the time in the \single task as well as in New York for the \fixedswitcher task, this number drops significantly when considering the \fulltask task.
Below we discuss key findings and patterns that warrant further analysis.

\paragraph{\noinstr and Waypoint Signalling}
Even though the \noinstr agent has no access to instructions, it performs surprisingly well; on some tasks even on par with other agents. Detailed analysis of the agent behaviour shows that it achieves this performance through a clever strategy: It changes the current direction based on the previous reward given to the agent (signalling a waypoint has been reached). Next, it circles around at the nearest intersection, determines the direction of traffic, and turns into the valid direction. Since many streets in New York are one-way, this strategy works surprisingly well. However, when trained and evaluated without the access to waypoint signal, it fails as expected (\nosignal in Table \ref{tab:singletask}, \noinstr in Table \ref{tab:hardtask}).

\paragraph{Non- vs. Attentional Architectures}
As expected, in the \single task performance of non-attentional agents is on par with
the attentional ones
(Table \ref{tab:singletask}). However, in \fixedswitcher the \enhancedvanilla agent has the upper hand over other models (Table \ref{tab:conetask}). Unlike \hafs, this agent can simultaneously read all available instructions, however, at the cost of possessing a larger number of weights and lack of generalisation to a different number of instructions.
That is, \hafs and \summationvanilla have roughly $N$ times fewer parameters than \enhancedvanilla, where $N$ is the number of instructions.

We observe that the \safs agent quite closely mirrors the performance of the \summationvanilla agent, and indeed the attention weights suggest that the agent pursues a similar strategy by mixing together all available instructions at a time. Therefore we drop this architecture from further consideration.

As an architectural choice, \enhancedvanilla is limited. Unlike the other models it cannot generalise to a larger number of instructions (Table \ref{tab:larger}). The same agent also fails on the hardest task, \fulltask (Table \ref{tab:hardtask}).

On the \fulltask, where reward is only available at the goal, \hafs outperforms the other models, mirroring its superior performance when increasing the number of instructions. This underlines our motivation for this architecture in decoupling the instruction following from the instruction selection aspect of the problem.

\paragraph{Supervised vs RL agents}
While the majority of our agents use RL, \student and \teacher are trained with a dense signal of supervision (Section \ref{sec:baselines_ablations}). The results in Table \ref{tab:singletask} show that the supervised agents can fit the training distribution, but end up generalising poorly. We attribute this lack of robustness to the highly supervised nature of their training, where the agents are never explicitly exposed to the consequences of their actions during training and hence never deviate from the gold path.
Moreover, the signal of supervision for \student turns the agent whenever it makes a mistake, and in our setting each error is catastrophic.

\begin{figure}
\centering
\includegraphics[width=0.8\linewidth]{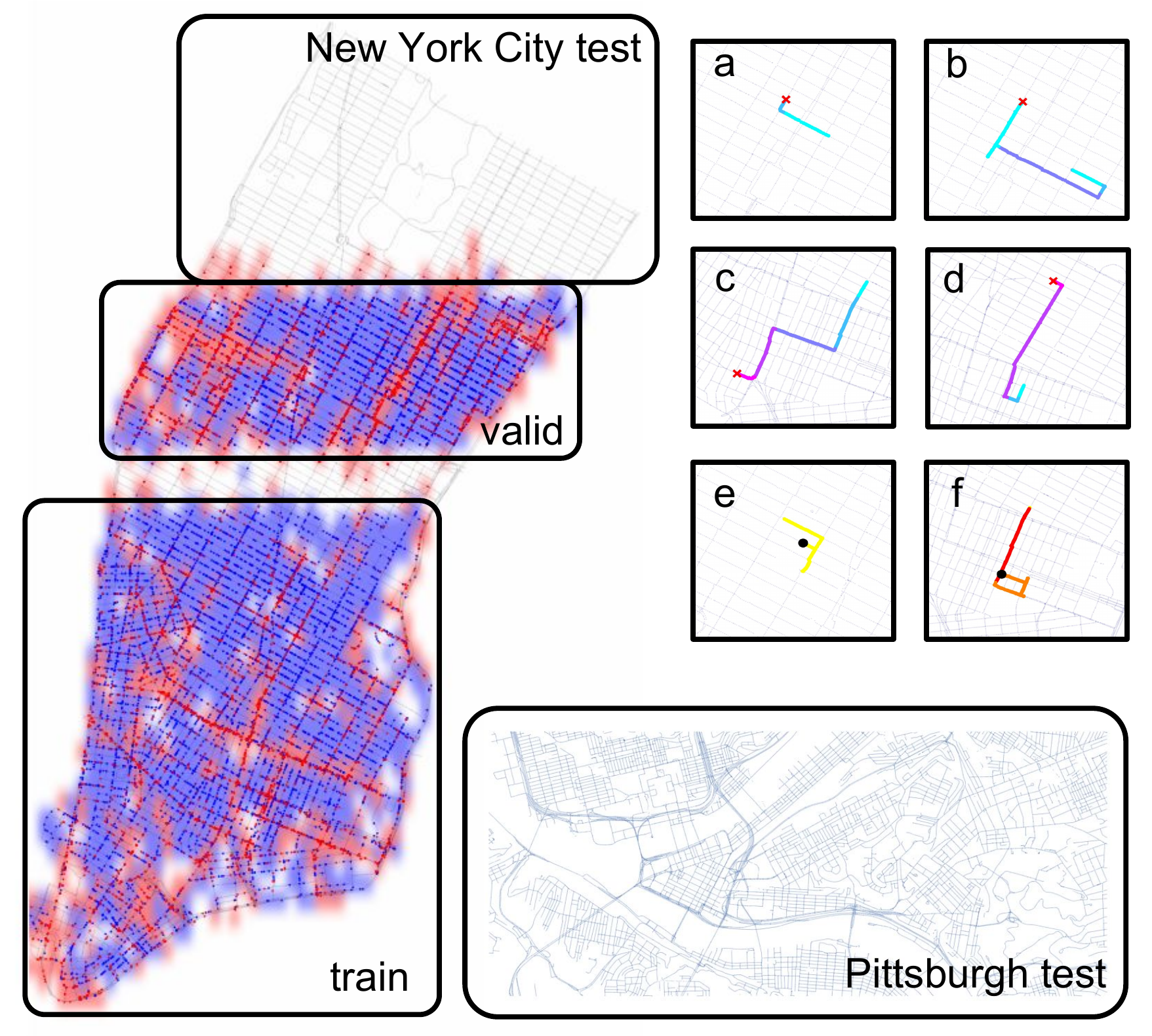}
\caption{Left: Map of Manhattan with training, validation and test areas, overlaid with the heat map of goal locations reached (blue) or missed (red) on training and validation data, using a \hafs agent trained on the \rawfixedswitcher task. Bottom right: Pittsburgh area used for testing. Top right: trajectories with color-coded attention index predicted by a \hafs agent with a learned switcher and trained on the \rawfulltask task; we show successful trajectories on validation (a and b) and on training (c and d) data, as well as two trajectories with missed goal (e and f).}
\label{fig:heatmap_nyc}
\end{figure}

\paragraph{Transfer}
We evaluate the transfer capabilities of our agents in a number of ways. First, by training, validating and testing the agents in different parts of Manhattan, as well as testing the agents in Pittsburgh. The RL agents all transfer reasonably well within Manhattan and to a lesser extent to Pittsburgh. The drop in performance there can mostly be attributed both to different visual features across the two cities and a more complex map (see Figure \ref{fig:heatmap_nyc}).

As discussed earlier, we also investigated the performance of our agents on a task with longer lists of directions than observed during training (Table \ref{tab:larger}). The declining numbers highlight the cost of error propagation as a function of the number of directions.

\paragraph{NoText and NoThumb}

As agents have access to both thumbnail images and written instructions to guide them to their goal, it is interesting to compare the performance of two agents that only use either one of these two inputs. The \nothumb agent consistently performs better than the \notext agent, which suggests that language is the key component of the instructions. Also note how \nothumb, which is based on the \enhancedvanilla architecture, effectively matches that agent's performance across all tasks, suggesting that thumbnails can largely be ignored for the success of the agents.
\notext outperforms the directionless baseline (\noinstr), meaning that the thumbnails by themselves also carry some valuable information.

\section{Conclusions and Future Work}

Generalisation poses a critical challenge to deep RL approaches to navigation, but we believe that the common medium of language can act as a bridge to enable strong transfer in novel environments. We have presented a new language grounding and navigation task that uses realistic images of real places, together with real (if not natural) language-based directions for this purpose.

Aside from the \streetlang environment and related analyses, we have proposed a number of models that can act as strong baselines on the task.
The hard-attention mechanism that we have employed is just one instantiation of a more general idea which can be further explored.
Other natural extensions to the models presented here include adding OCR to the vision module and developing more structured representations of the agents' state. Given the gap between reported and desired agent performance here, we acknowledge that much work is still left to be done.

\bibliographystyle{aaai}
\fontsize{9.0pt}{10.0pt}
\bibliography{bibliography}

\end{document}